\title{Information-Theoretic Scoring Rules to Learn Additive Bayesian Network Applied to Epidemiology}

\author{
        Gilles Kratzer \\
        Department of Mathematics \\
        University of Zurich\\ 
        Zurich, Switzerland %\underline{Switzerland} 
            \and
        Reinhard Furrer\\
        Department of Mathematics \\
        Department of Computational Science\\
        University of Zurich\\
        Zurich, Switzerland %\underline{Switzerland}
        }
\date{\today} 

\providecommand{\keywords}[1]{\textbf{\textit{Keywords---}} #1}

\documentclass[12pt]{article}

\usepackage[square,sort,comma,numbers]{natbib}
\usepackage{amsmath}
\usepackage{algorithm2e}
\usepackage{graphicx}
\usepackage{float}
\usepackage{amssymb}
\usepackage{caption}
\usepackage{subcaption}
\usepackage{fancyvrb}

\newcommand{\Abn}{{\mathcal{A}}}

\newcommand{\Cmp}{{\mathcal{C}}}
\newcommand{\Data}{{\mathcal{D}}}

\newcommand{\Par}{{\mathbf{Pa}_j}}
\newcommand{\Prob}{{\mathcal{P}}}

\providecommand{\rr}[1]{{\small\tt #1}}
 
\begin{document}

\maketitle
 
\begin{abstract}
Bayesian network modelling is a well adapted approach to study messy and highly correlated datasets which are very common in, e.g., systems epidemiology. A popular approach to learn a Bayesian network from an observational datasets is to identify the maximum a posteriori network in a search-and-score approach. Many scores have been proposed both Bayesian or frequentist based. In an applied perspective, a suitable approach would allow multiple distributions for the data and is robust enough to run autonomously. A promising framework to compute scores are generalized linear models. Indeed, there exists fast algorithms for estimation and many tailored solutions to common epidemiological issues. The purpose of this paper is to present an R package {\sf abn} that has an implementation of multiple frequentist scores and some realistic simulations that show its usability and performance.  It includes features to deal efficiently with data separation and adjustment which are very common in systems epidemiology.
\end{abstract}


\section{Introduction and background}\label{intro} 

Bayesian Network (BN) modelling has an impressive track record in analysing systems epidemiology datasets \cite{cornet2016bayesian, mccormick2014frequent}, especially in veterinary epidemiology \cite{mccormick2013using,ludwig2013identifying, firestone2014applying,cornet2016bayesian, RUCHTI2018}. It is a particularly well suited approach at the beginning of the coil of discovery within a field of research to better understand the underlying structure of the data. It is designed to sort out directly from indirectly related variables and offers a far richer modelling framework than classical approaches in epidemiology, like, e.g., regression techniques or extensions thereof. In contrast to structural equation modelling \cite{hair1998multivariate}, which requires expert knowledge to design the model, the Additive Bayesian Network (ABN) method is a data-driven approach \cite{lewis2013improving, kratzerabn}. It does not rely on expert knowledge but it can possibly incorporate it.  

\subsection{Previous work and other implementations}\label{previous_work}
 
There exists two broad classes of algorithms to learn BNs. The constraint-based approaches, where one learns the BN using statistical independence tests. The optimal network is identified using the reciprocal relationship between graphical separation and conditional independence \cite{spirtes2001anytime}. An other popular approach is based on network scoring. The idea is that each candidate network is scored and the one which has the largest score is kept. In practice this is computationally intractable for a typical number of variables. Indeed, the number of possible networks is massive and increases super-exponentially with the number of nodes \cite{robinson1977counting}. A practical workaround is to use a decomposable score, i.e., a score that is additive in terms of network's node and depends only on the parents of the index node. This approach is very close to the classical model selection in statistics \cite{zou2017model}. The scoring approach paradigm is that the scores should represents how well the structure fits the data \cite{zou2017model}. Many scores have been proposed for discrete BNs in a Bayesian context \cite{daly2011learning} that aims at maximising the posterior probability. Indeed \citet{heckerman1995learning} proposes the so-called the Bayesian Dirichlet (BD) family of scores. It regroups the K2, BDeu, BDs, BDla scores \cite{scutari2018dirichlet}. 

On the other hand, scores within a frequentist framework have been proposed \cite{daly2011learning}, such as Bayesian Information Criterion (BIC), Akaike Information Criterion (AIC), Minimum Description Length (MDL) \cite{daly2011learning}. They all have in common a goodness-of-fit part and a penalty for model complexity. Indeed, the naive idea to use the maximum likelihood as a score is suboptimal as it generally selects a fully connected graph. 

One major limitation of the scoring approach is that many different networks can have the same score. Networks which have the same scores are in the same equivalence class. The BDeu is the only score-equivalent BD score. BDs is only asymptotically score equivalent. One interesting feature is that BIC is also score equivalent for discrete BNs. In a causal perspective, i.e., when arcs direction matters, scores which are equivalent are preferred \cite{scutari2018dirichlet}.

\medskip

R as an open source, reliable and easy to use environment for statistical computing is very popular in the epidemiological community \cite{R-Core-Team:2017aa}. One popular R package for BN learning is the package {\sc bnlearn} \cite{Scutari:2010aa}. It has implementations of most of BD's scores but also AIC and BIC for continuous and discrete mixed variables.  Additionally, it has implementations of multiple network structure learning via multiple constraint-based and score-based algorithms. Focused on a causal framework, {\sc pcalg} is very popular \cite{Kalisch:2012aa}. It has an implementation of the PC-Algorithm that selects one class representative of the network skeleton. Other useful R packages are actively maintained on CRAN. But none of them has an implementation of scoring procedure that deals simultaneously for multiple exponential family representative in a likelihood based framework. Beside R, it exists multiple implementation in other computing environment such as Weka \cite{bouckaert2008bayesian}, Matlab \cite{murphy2001bayes} or open sources python or C++ implementation. But in the epidemiological community the R implementation are the most popular ones due to the simplicity of use of~R.

\subsection{Motivations for frequentist scoring}

One actual strong limitation of the R package {\sc abn} is that it cannot deal with multinomial distributed random variables. Indeed, in applied epidemiology, the data are often composed of a mixture of different distributions. The most common are categorical, continuous and discrete random variable. The R package {\sc abn} has an implementation of binomial, Gaussian and Poisson Bayesian regression. This implies that when dealing with categorical random variable one has to split the random variable in a pairs. This is interesting from a modelling perspective as this is very flexible. But it destroys the intrinsic link between the dichotomised variables which can negatively affects the modelling. Additionally, it increases the number of variables which is computationally not desirable.

A surprisingly common problem when dealing with binomial regression in an applied perspective is the so-called data separation problem \cite{gelman2008weakly}. This arises when a covariate predicts perfectly the outcome. A related issue is the quasi data separation (or data sparsity), when the number of observation per class is dramatically low. In such configuration, the maximum likelihood estimates tends to diverge and the classical algorithms for estimation becomes numerically unstable. A possible workaround is to exclude predictors that create separation. This is in general a suboptimal idea as those predictors are specifically the ones that are the most promising to explain the outcome \cite{zorn2005solution}. Related approaches would be to exclude specifically cases that are separated, to collapse predictors categories or to bin predictor values. In a machine learning perspective when dealing with a large number of variables and automatised procedures require robustness. Those data tailored workarounds scale laboriously. An other option is to change a few randomly selected observations in order to render the problem estimable. This implies to see separation as a problem per se rather than as a symptom of paucity of information in the data. 

In a Bayesian framework, a possible solution is to use the prior to drive the posterior when the likelihood fails to correctly estimate the information contained in the data. This implies to put some information in the prior. In a frequentist approach, one can penalise the likelihood in order to prevent from becoming infinite. This approach, called bias reduction, can be easily implemented in an iterative scheme by generating pseudo data \cite{kosmidis2009bias}. Surprisingly, this latter approach originally proposed by \citet{firth1993bias} is equivalent to penalised the log-likelihood using a Jeffreys prior \cite{kosmidis2009bias}. Due to the simplicity of implementation the bias-reduced estimation was the chosen solution for data separation in {\sc abn}.

In epidemiology, statistical adjustment is often used to control for the effect of supposed confounding factors and make comparison according to different populations more easy \cite{wilcosky1985comparison}. {\sc abn} has an option to compute statistically adjusted structures. The adjusting variables are imposed as covariate to every models computed in the cache. 

\section{Additive Bayesian networks}

ABN models are graphical models that extend the usual Generalized Linear Model (GLM) to multiple dependent variables through the factorisation of their joint probability distribution \cite{lewis2013improving}. They are represented with a Directed Acyclic Graph (DAG) and the set of parameter estimates. An ABN model assumes that each node is a GLM where the covariate are the parents and the distribution depends on the index node. The model learning phase is a two steps process: 1. Network, skeleton or structure learning ($\mathcal{S}$) 2. Parameter learning, the model parameter are $\theta_{\mathcal{M}}$. Hence in a Bayesian framework, constructing an ABN model $\mathcal{M}$ given a set of data, $\mathcal{D}$ is:
\begin{align*}
& P(\mathcal{M}\mid \mathcal{D}) = \underbrace{\,P( \theta_\mathcal{M} , \mathcal{S}\mid\mathcal{D})\,}_{\text{model learning}} =
\underbrace{\,P(\theta_\mathcal{M}\mid\mathcal{S},\mathcal{D})\,}_{\text{parameter learning}} ~ \cdot  \underbrace{\,P(\mathcal{S}\mid\mathcal{D})\,}_{\text{structure learning}}.
\end{align*}

The two learning steps are interconnected. Several efficient algorithms have been proposed for both learning procedures. In order to learn the relationships between variables, the conditional probability distributions tables should be learned, this can be done in a frequentist setting using the classical Iterative Reweighed Least Square (IRLS) algorithm \cite{faraway2016extending}. The structure selection step can be done using a heuristic or exact approach.

An interesting feature of the ABN methodology is to be able to impose external expert knowledge. Indeed, in most of applied data analysis, some part of the network is known. For example, if two random variables are timely related the direction of the possible arrow is known. Or if, based on existing literature a possible connection is known to be expected. {\sc abn} allows such external causal input through a banning or a retaining matrix. Those matrices are used to compute the list of valid parent combination. However, a more theoretically sounding approach, suggested by \citet{heckerman1995learning}, is to use augment the observed data with synthetic data that represents the causal belief. This approach is superior as it solves the problem of the likelihood equivalence posed by banning and retaining part of the structure, but its feasibility in an ABN analysis remains an open question \cite{mccormick2013using}.

\subsection{Scoring procedure}

The crucial points regarding BN learning process is speed and resilience. Indeed, {\sc abn} computes autonomously a comprehensive cache of licit scores. More precisely it computes all allowed combinations, i.e. not banned, of a node and his set of parents up to a given complexity ideally without end user involvment.

%A fast algorithm for the estimation of GLM parameters is an iterative scheme. Classically a statistical lecture would present GLM framework as a non linear model of the form:
%
%\begin{equation}
%g(Ax) =b + e ,  \qquad \qquad V(e)= \sigma W^{-1}
%\end{equation}
%
%Where $A$ is m $\times$ n matrix of rank n, $e$ is a vector random variable with covariance matrix $\sigma W^{-1}$.  $W$ is a symmetric positive definite matrix. $g()$ is the link function that aims at constrain GLM solution to a given support. See \citet{faraway2016extending} for a detailed introduction to GLM. The minimum squared solution is obtained in solving: 
%
%\begin{equation}
%\min_{x} g(Ax)-b)^{\top}W(g(Ax)-b
%\end{equation}
%
A popular choice for estimating GLMs is the IRLS algorithm \cite{faraway2016extending}. However, IRLS implementations are known to be potentially numerically unstable, mainly due to rank deficiency and we will treated this issue separately. The IRLS algorithm is able to estimate every distribution of the exponential family. The distributions of interest for {\sc abn} are: Gaussian, binomial and Poisson. The multi-categorical random variables can be estimated through IRLS like algorithm. But a well known weakness of this approach is the very high per-iteration cost due to sparsity of the intermediate matrices. However, fast estimation of multinomial logit models that efficiently take advantage of the matrix structures have been proposed \cite{hasan2014fast}. It overshoots the purpose of {\sc abn} as it requires tailored parametrisation. An alternative approach to estimate multinomial logistic regressions is to approximate it using multiple sequential binary logistic regressions. The primary purpose of the scoring procedure is to estimate goodness-of-fit metrics, then special care should be taken to compute the log-likelihood as each observations are counted multiple times. This estimation step substantially slows down the estimation process.  A robust, easy to implement and fast solution to estimate unregularized multinomial regression models is to use  neural networks with no hidden layers, no bias nodes and a softmax output layer. The optimisation is done through maximum conditional likelihood. This procedure is implemented in R by the {\sc nnet} package \cite{Venables:2002aa}.

\subsection{Information theoretic scores}

The implemented scores in {\sc abn} are AIC, BIC and MDL:
\begin{align}
AIC(\Abn_s, \Data) &= - \log \Prob(\Data | \hat{\Theta}, \Abn_s) + 2 d,\\
BIC(\Abn_s, \Data) &= - \log \Prob(\Data | \hat{\Theta}, \Abn_s) + d/2 \log n,\\
MDL(\Abn_s, \Data) &= - \log \Prob(\Data | \hat{\Theta}, \Abn_s) + d/2 \log n + \Cmp_k,
\end{align}
where $\hat{\Theta}$ is the maximum likelihood parameters for $\Abn_s$ and $ \Prob(\Data | \hat{\Theta}, \Abn_s)$ is the maximum value of the likelihood function, $d$ is the number of free parameter, $n$ is the sample size,  $k$ is the number of variables and  $\Cmp_k = \sum_{j=1}^{k} (1+ |\Par |) \log k$. $ |\Par |$ is the size of the parent set of variable $j$. Then the MDL is the BIC with an extra penalty for model complexity.

\section{Software implementation}

An ABN analysis is done using successive application of three functions.

\begin{Verbatim}[fontsize=\small]
bsc.compute <- buildscorecache.mle(data.df = df.abn., 
                                data.dists = dist, max.parents = 5)
dag <- mostprobable(score.cache = bsc.compute, score = "bic")
fit.dag <- fitabn.mle(dag.m = dag, data.df = bsc.compute$data.df, 
                      data.dists = dist)
\end{Verbatim}

In the R package {\sc abn} the \rr{buildscorecache.mle()} function is essentially a wrapper of the \rr{fitabn.mle()} function. It requires minimally: a named dataset, the named list of the distribution of the entries of the dataset and an upper limit for network complexity. It computes firstly an empty list of valid parent combination, using banning and retaining input matrices (which are assumed to be empty by default). Then it iterates through the cache to score the candidate piece of network. At each step of the scoring the used IRLS algorithm depends on the given list of distributions. For the special case of the binomial nodes, the usual logistic regression is tried, if it fails to estimate the given problem,  a bias-reduced tailored algorithm is used. If however the algorithm fails to return a finite estimate, some predictors are sequentially removed until the design matrix becomes fully ranked. These three steps ensure {\sc abn} to be able to score a dataset even if it is data separated. In this example, an exact search algorithm is used to select the maximum a posteriori DAG (\rr{mostprobable()}) \citet{koivisto2004exact}.

The \rr{fitabn.mle()} function scores a given network. It requires a valid DAG, a named datasets and a named list of distributions. It returns the list of score for each node, the parameter estimates, the standard deviation and the $p$-values. Special care should be taken when interpreting and displaying the $p$-values. Indeed, the DAG has been selected using goodness of fit metric so adjustment methods should at least be used. 

The Bayesian equivalent functions in {\sc abn} are \rr{fitabn()} and \rr{buildscorecache()}. Those function estimate Bayesian regression using following parameter priors: weekly informative Gaussian priors with mean zero and variance 1000 for each of the regression parameters of the model (both binomial and Gaussian), as well as diffuse Gamma priors (with shape and scale equal to 0.001) for the precision parameters in Gaussian nodes in the model. 

%\subsection{Pseudocode}

\subsection{Benchmarking}

In order to benchmark the performances of the different possible implementations, complex BNs of 10 purely continuous nodes have been simulated. Then a dataset of 10'000 observations have been simulated from each BN. The benchmarking is performed based on 50 repetitions \cite{Mersmann:2018aa}. The different algorithms are:
\begin{enumerate}
\item fitabn: is the fitting procedure based on a tailored INLA code written in C (available in abn) \cite{lewis2013improving};
\item glm.irls\_cpp: is the Rcpp implementation of the IRLS algorithm (described in this article);
\item glm.irls\_qrnewton: is an R implementation of the IRLS algorithm based on a QR decomposition in a Newton scheme;
\item glm.irls: is the R implementation of the IRLS algorithm \cite{R-Core-Team:2017aa};
\item glm.speedglm.wfit: is the R function \rr{speedglm.wfit()} well adapted for matrix from {\sc speedglm} in an switch scheme \cite{Enea:2017aa};
\item glm.speedglm: is the R function \rr{speedglm()} from {\sc speedglm} in a switch scheme \cite{Enea:2017aa};
\item glm.switch: is the R function \rr{glm()} from {\sc stats} in a switch scheme \cite{R-Core-Team:2017aa}.
\end{enumerate}

\begin{figure}[!h]
  \centering
\includegraphics[width=\textwidth]{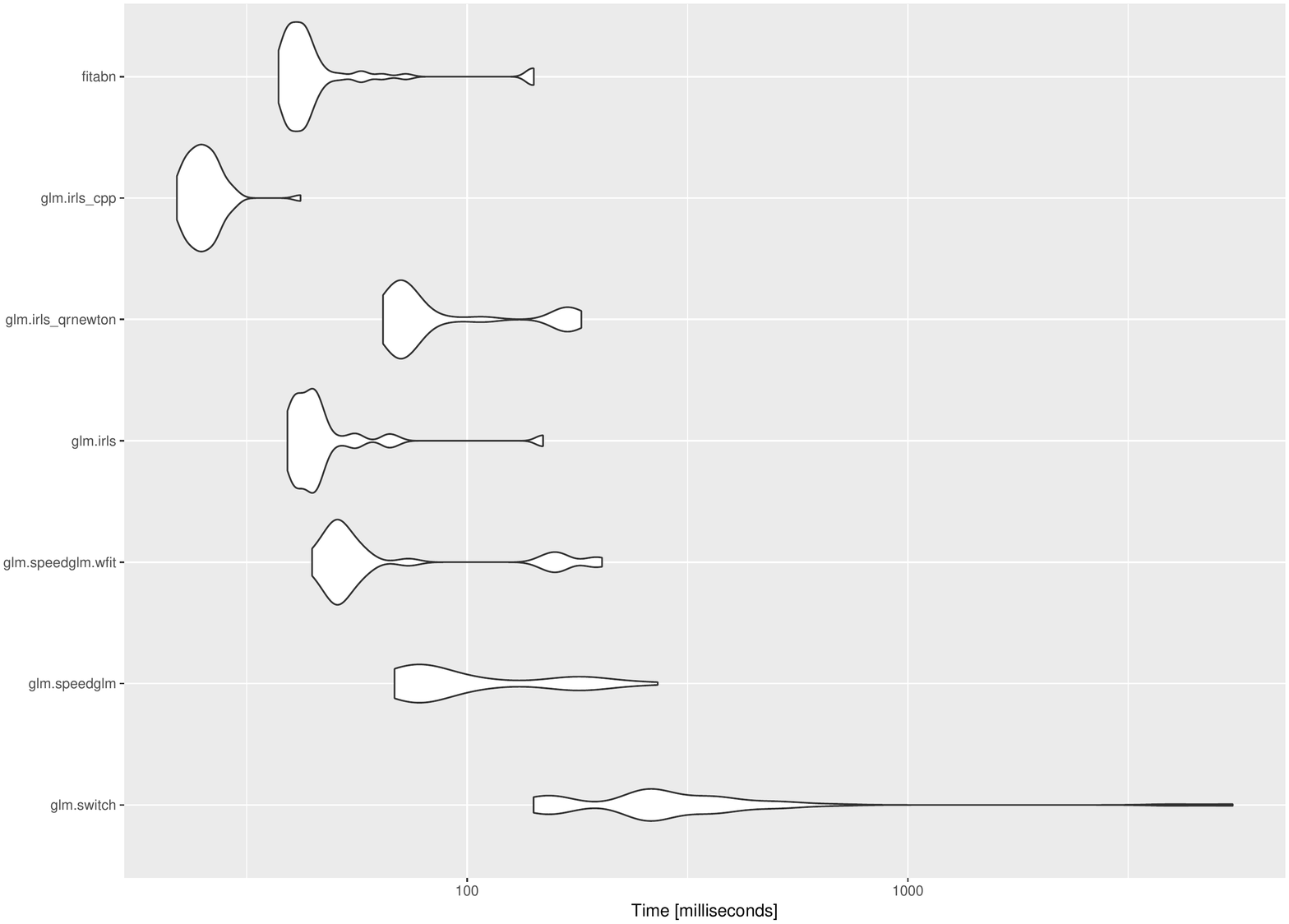}
  \caption{Benchmarking comparison between different algorithm implementations to estimate regression coefficient in an abn framework on a logarithmic scale.}
  \label{fig:bench}
\end{figure}

As one can see in Figure~\ref{fig:bench}, the Rcpp implementation (glm.irls\_cpp) is the fastest which was one of the motivations of this project. However, the comparison is somewhat unfair as the different functions do not return the same output. 
 
\section{Simulation studies} 
 
In order to assess the performance of the {\sc abn} implementation a simulation study have been performed. The parameter which are important from a simulation point of view are: the BN dimension (i.e. the number of nodes of the BN), the structure density (i.e. average number of parent per node) and the sample size. Additionally to those structure-wise metrics important factors impacting simulations are the intensity of the arc link, the variability of the arcs distributions and the mixture of variables. Indeed, scores used are only approximately score independent and increases in mixing distribution. Simulating randomly DAG structures are done through the simulation of adjacency matrices with constrains for ensuring acyclicity. Simulating observations from a given structure is done with random number generator in respecting the node ordering using JAGS \cite{plummer2003jags}. 

\subsection{Regression coefficients estimation}

\begin{figure}[!h]
  \centering
\includegraphics[width=\textwidth]{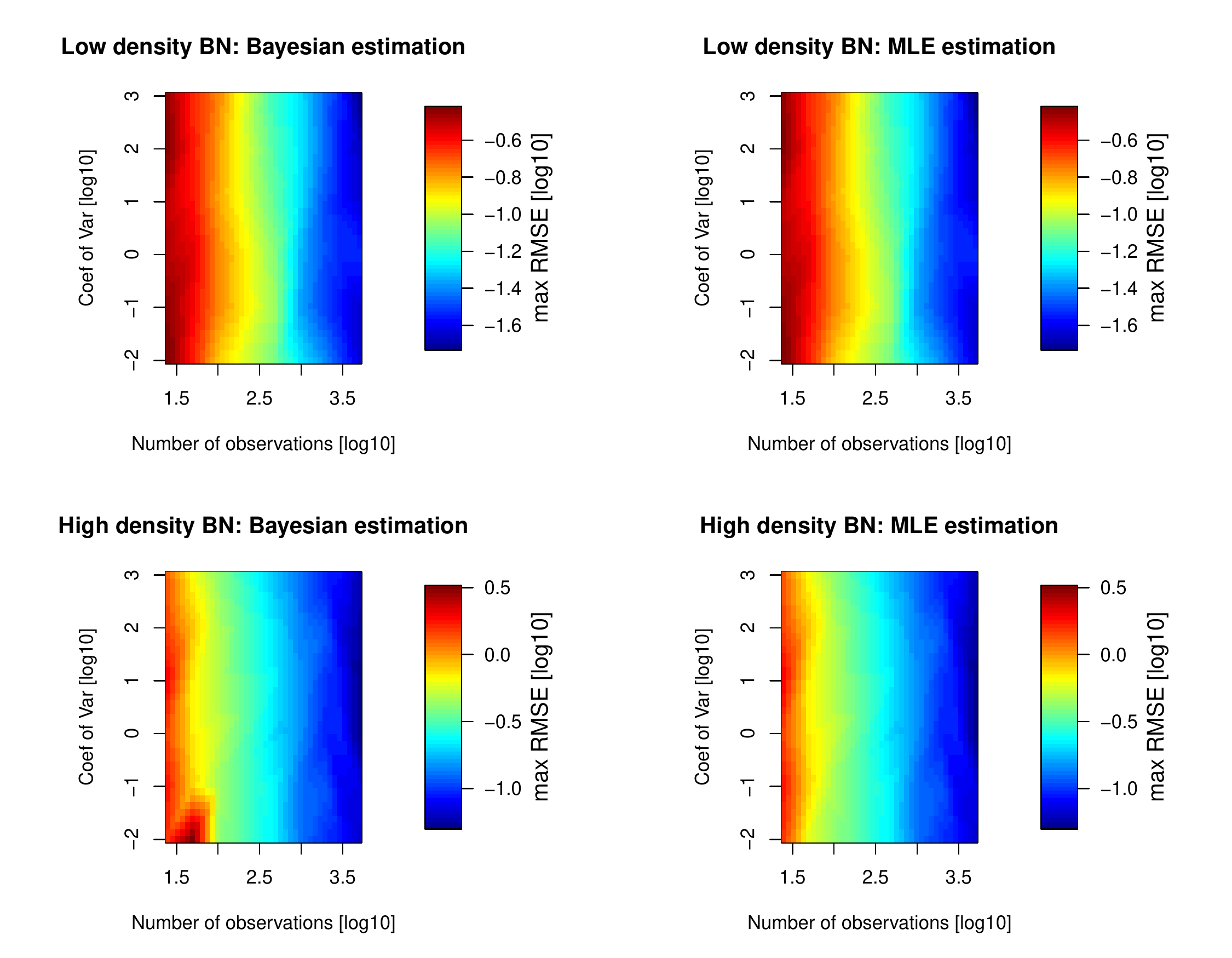}
  \caption{Comparison between Bayesian and MLE implementation to estimate regression coefficient in an ABN framework. The panels show the maximum Root Mean Squared Error (max RMSE) in function of the network density, the distribution variability and the sample size.}
  \label{fig:arc_estimation}
\end{figure}

The Bayesian and MLE implementation are compared for parameter estimation accuracy in Figure~\ref{fig:arc_estimation}. Two network densities, 20\% and 80\% of the possible arcs expressed, have been simulated 50 times. Then the regression coefficients have been computed for different sample size and the coefficient of variation of the given node as a proxy for distribution variability. 

As one can see in Figure~\ref{fig:arc_estimation}, the error measured as the maximum root mean squared error (max RMSE) on a log-log scale, both implementations produce very similar results. Even if the estimation frameworks are very different. 

\subsection{Score efficiency comparison}

In order to estimate the efficiency of the different scores implemented in {\sc abn}. BNs with a given arc density have been simulated. Form those networks 20 datasets have been simulated for different sample sizes. The metrics used to display performance of the score are: the true positive (number of arc retrieved), false positive (learning an arc where there is not) and false negative (learning no arc where there is one). The scores are used are abn (marginal posterior likelihood in a Bayesian regression framework), mlik (maximum likelihood) AIC, BIC and MDL.

\begin{figure}[!h] 
  \centering
\includegraphics[width=\textwidth]{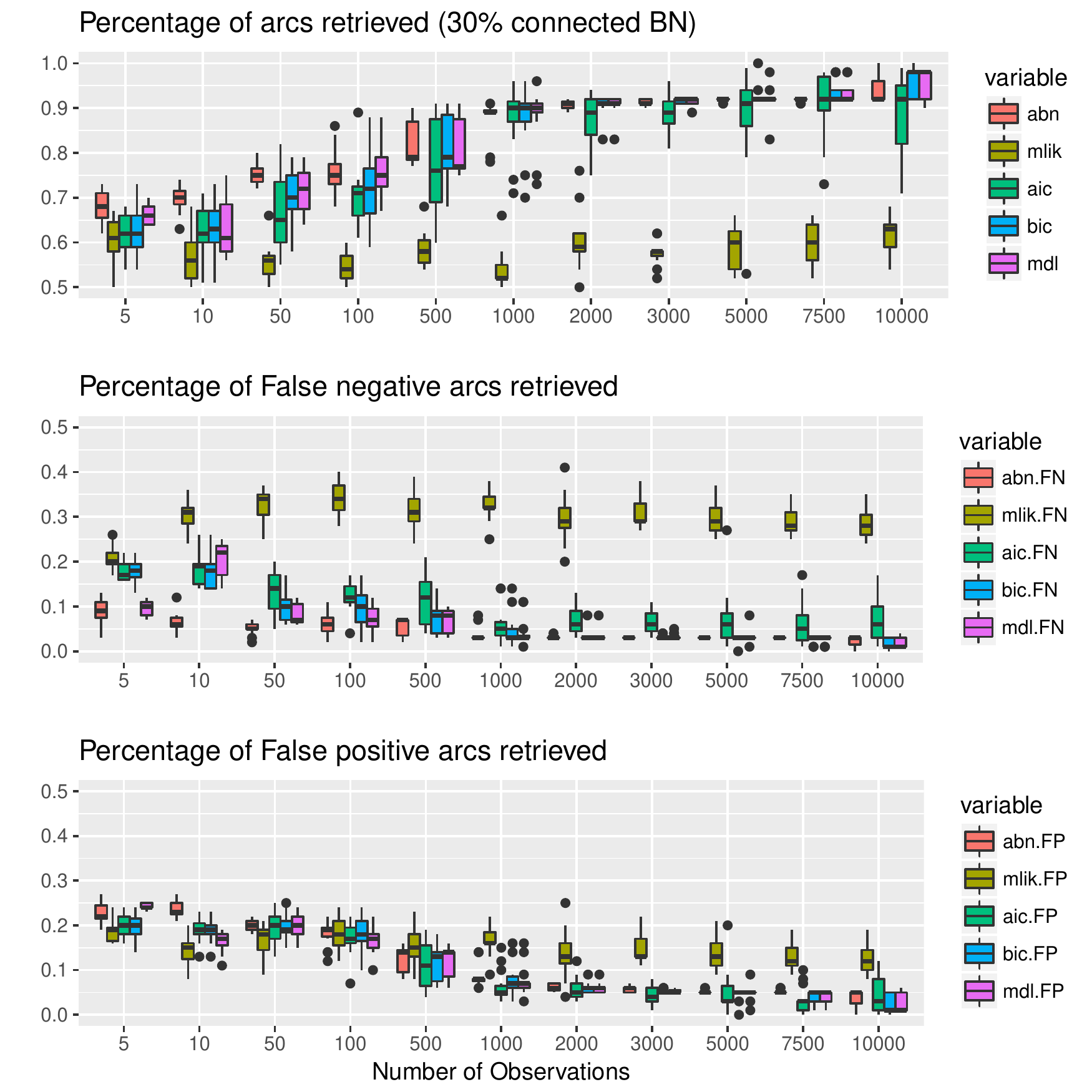}
  \caption{Score efficiency comparison for abn (marginal posterior likelihood), mlik (maximum likelihood), AIC, BIC and MDL for a given BN density in function of the number of observations using true positive, false positive and false negative metrics.}
  \label{fig:score}
\end{figure}

As one can see in Figure~\ref{fig:score} the mlik (maximum likelihood) is a suboptimal score for BN learning as expected by \citet{daly2011learning}. Indeed it tends to return fully connected BN. The Bayesian score (abn) seems to be the most efficient especially when the sample size is low. This score seems to be especially better at identifying the absence of arcs compared to other scores. The AIC seems to perform less effectively than the other information-theoretic scores due to its larger variability.

%\subsection{ROC analysis}
%
%\section{Motivating example}
%\section{Conclusion and impact}

\clearpage

\bibliographystyle{abbrv}

\bibliography{bib_abn}

\end{document}